\documentclass[11pt,letterpaper]{article}
\usepackage{emnlp2017}
\usepackage{times}
\usepackage{latexsym}
\usepackage{url}

\usepackage{amsmath}
\usepackage{amssymb}
\usepackage{bm}
\usepackage[ruled,linesnumbered]{algorithm2e}
\usepackage{multirow}
\usepackage[font={normalsize}]{caption}
\usepackage{graphicx}

\usepackage{float}
\floatstyle{plaintop}
\restylefloat{table}

\newtheorem{definition}{Definition}

\emnlpfinalcopy

\def\emnlppaperid{***}

\title{Supervised Complementary Entity Recognition \\with Augmented Key-value Pairs of Knowledge}

\author{Hu Xu$^1$, Lei Shu$^1$, \and Philip S. Yu$^{1, 2}$\\
    $^1$Department of Computer Science, University of Illinois at Chicago, USA\\
    $^2$Institute for Data Science, Tsinghua University, Beijing, China\\
    \{hxu48, lshu3, psyu\}@uic.edu\\
}

\date{}

\begin{document}

\maketitle

\begin{abstract}
Extracting opinion targets is an important task in sentiment analysis on product reviews and complementary entities (products) are one important type of opinion targets that may work together with the reviewed product. In this paper, we address the problem of Complementary Entity Recognition (CER) as a supervised sequence labeling with the capability of expanding domain knowledge as key-value pairs from unlabeled reviews, by automatically learning and enhancing knowledge-based features. We use Conditional Random Field (CRF) as the base learner and augment CRF with knowledge-based features (called the Knowledge-based CRF or KCRF for short). We conduct experiments to show that KCRF effectively improves the performance of supervised CER task.
\end{abstract}

\section{Introduction}
\label{sec:intro}
Aspect extraction (or opinion target extraction) is an important task in sentiment analysis \cite{pang2002thumbs} on product reviews \cite{hu2004mining,popescu2007extracting,liu2015sentiment}. Besides opinion targets about the reviewed product itself (e.g., the ``screen'' in ``iPhone's screen is great.''), products that are compatible or incompatible with the reviewed product are also important opinion targets. Extracting those kinds of opinion targets are studied as Complementary Entity Recognition (CER) in \cite{xu2016CER,xu2016mining}. For example, if the sold product is a \textit{tablet stand}, then the ``iPhone'' in ``This tablet stand works with my iPhone'' is a complementary entity. Note that CER highly relies on entity's contextual information (e.g., the word ``works'' in the previous example) and such information can be domain dependent. For example, ``holds'' in ``This tablet stand holds my iPhone well'' is a context verb particular for \textit{tablet stand}. A traditional supervised method like Conditional Random Field (CRF) \cite{lafferty2001conditional} may have good precision on such extraction yet suffer from low recall due to unseen context words appear in the test data but not in the training data. To solve this problem, instead of using supervised method, \cite{xu2016CER} uses an unsupervised method by leveraging manually-crafted high precision dependency rules \cite{bach2007review,culotta2004dependency,bunescu2005shortest,joshi2009generalizing} to expand (bootstrap) context words as knowledge on a large amount of unlabeled data and combine those context words with another set of manually-crafted high recall dependency rules for CER. However, crafting dependency rules for both context words and CER can be time-consuming and such rules may be domain dependent and subject to change for new domains.

To benefit from both the supervised and unsupervised methods, we consider to automatically learn patterns for both CER and knowledge expansion (of context words) from training data and expand context word knowledge on unlabeled data. So when making predictions on the test data, the model can leverage more contextual knowledge from unlabeled data to make better predictions. Or put it another way, we wish the prediction behavior of a supervised model can change after training when it sees more unlabeled data. This framework is inspired by the lifelong sequence labeling method proposed in \cite{shu2017lifelong,shu2016supervised}. However, we do not expand knowledge for lifelong learning here and we make one step further: we automatically learn knowledge-based features and knowledge (or context words) as key-value pairs rather than manually crafting them. We use CRF as the base learner and augment CRF with knowledge-based features (a modified dependency relation) that are automatically learned from the training data. The augmented CRF is called Knowledge-based CRF (KCRF).

The proposed method has the following steps:\\
\textbf{Pre-training} We first train a CRF as a traditional sequence labeling training process using hand-crafted features, including primitive features (defined later) such as dependency relation based features. Then we automatically select from those primitive features as knowledge-based features to build a group of key-value pairs as initial knowledge, where keys are selected feature types and values are feature values (e.g., context words) appear in the training data.\\ 
\textbf{Knowledge-based Training} Then we train a Knowledge-based CRF (KCRF) based on the initial knowledge. So KCRF knows which features (as keys) can be used to expand knowledge (get more values for the same key).\\
\textbf{Knowledge Expansion} We expand the values in initial knowledge by iteratively collecting reliable knowledge from reliable predictions on plenty of unlabeled reviews. Experiments demonstrate that the expanded knowledge is effective in predicting test data.

\section{Preliminaries}
We briefly review the terms used throughout this paper. We use dependency relations as the major type of knowledge-based features since a dependency relation associates one word (current word) with another word (context word), which can be viewed as a piece of context knowledge.

\begin{definition} [Dependency Relation] \label{defn:dr} 
A \emph{dependency relation} is a typed relation between two words in a sentence with the following format: 
$$\textit{(type, gov, govpos, dep, deppos)}, $$
where \emph{type} is the type of a dependency relation, \emph{gov} is the \emph{governor word}, \emph{govpos} is the POS (Part-Of-Speech) tag of the governor word, \emph{dep} is the \emph{dependent word} and \emph{deppos} is the POS tag of the dependent word. 
\end{definition}

\begin{definition} [Dependency Feature] \label{defn:df} 
A \emph{dependency feature} for the $n$-th word is a simplified dependency relation with the following attributes:
$$\textit{(role, type, gov/dep, govpos/deppos)}, $$
where \emph{role} can be either ``GOV'' or ``DEP'' indicating whether the $n$-th word is a governor word or a dependent word; \emph{type} is the type of the original dependency relation; \emph{gov/dep} is the other word associated with the $n$-th word via the original dependency relation and \emph{govpos/deppos} is the POS tag of the other word.
\end{definition}

Note here we omit the $n$-th word, its POS-tag in a dependency relation and define them as separate features since they are the same for all dependency features of the $n$-th word. 

\begin{definition} [Primitive Feature] \label{defn:pf} 
A \emph{primitive feature} can be either a dependency feature or a current word feature (taking current word as a feature). Primitive features are used to generate knowledge base.
\end{definition}

\begin{definition} [Knowledge Base] \label{defn:kb} 
A \emph{knowledge base} is a set of key-value pairs $(k, v) \in \textit{KB}$, where $k$ is the \emph{knowledge type} and $v$ is the \emph{knowledge value} belonging to that type. The same $k$ may have multiple knowledge values. We further define separate knowledge bases $\textit{KB}^{t_o}$ for each tag $t_o \in T$, where $T$ is the set of output labels for sequence labeing and $\textit{KB}=\{\textit{KB}^{t_o}|t_o \in T\}$.
\end{definition}

\begin{definition} [Knowledge-based Feature] \label{defn:kbf} 
A \emph{knowledge-based feature} is defined based on a knowledge type $k$ in a knowledge base. We use $d \in D$ to denote an index about a knowledge-based feature in a feature vector $x_n$. So $x_{n,d}=1$ indicates that the $d$-th feature of the $n$-th word is a knowledge-based feature of type $k_d$ with some $(k_d, v)$ found in \textit{KB}. We use $K=\cup_{t_o \in T} K^{t_o}$ to denote all knowledge types in \textit{KB}. 
\end{definition}

A primitive feature can generate a knowledge-based feature in the form of $(k, v)$. Current word feature has a corresponding knowledge type $k=\text{[WORD]}$ and takes the current word as the knowledge value $v$ (e.g., we use $(\text{[WORD]}, \text{``phone''})$ to indicate ``phone'' is in the knowledge base as type $\text{[WORD]}$ ). Dependency features have a corresponding knowledge type $k=\text{[role, type, govpos/deppos]}$, which is similar to a dependency feature. The \textit{gov/dep} part (the other word related to the current word in a dependency relation) is considered as the knowledge value $v=\text{gov/dep}$. For example, if \textit{K}=\{\text{[WORD]}, \text{[DEP, nmod:with, VBZ]}, \text{[GOV, nmod:poss, PRP\$]}\} and we have knowledge value \text{``phone''} and \text{``works''} for the first two types, we may have \textit{KB}=\{(\text{[WORD]}, \text{``phone''}), (\text{[DEP, nmod:with, VBZ]}, \text{``works''}) \}. We describe how to automatically obtain all knowledge types $K$ and how to get initial knowledge values in the next section.

\section{Pre-training}
The role of pre-training is to identify knowledge types $K$ and initial knowledge values. It is important to obtain useful knowledge types and reliable knowledge values because some knowledge types may not help the prediction task and wrong knowledge values may be harmful to the performance of predictions. Fortunately, a trained CRF model can tell us which features are more useful for prediction and need to be enhanced with knowledge. The basic idea is to perform a traditional CRF training using primitive features and select knowledge-based features $K$ and initial knowledge values based on the weights $\lambda$ of primitive features in the trained CRF model. 

Let $x_n'$ denote a feature vector of the $n$-th word in an input sequence for pre-training. We use $r \in R$ to denote an index about a primitive feature so $x'_{n,r}=1$ means the $n$-th word has a primitive feature (e.g., \textit{WORD=``phone''} or \textit{(DEP, nmod:with, works, VBZ)}) indexed by $r$. We distinguish different feature functions according to the value of $y_n$ and the primitive features indexed by $r$ in $x'_n$. We care about the following type of feature function, which is a multiplication of 2 indicator functions: 
\begin{equation} \label{eq:m}
\begin{split}
f_r^{t_o}(y_{n}, x'_n)=\mathbb{I}(y_{n} = t_o)\mathbb{I}(x'_{n,r} ) ,
\end{split}
\end{equation}
It turns all combinations of primitive features $r \in R$ and tag set $T$ into $\{0, 1\}$. Further we use a similar notation $\lambda_r^{t_o}$ for the corresponding weight. A positive weight $\lambda_r^{t_o}>0$ indicates a primitive feature indexed by $r$ has positive impact on predicting tag $t_o$; while a negative weight $\lambda_r^{t_o}<0$ indicates a primitive feature indexed by $r$ has negative impact on predicting tag $t_o$.

After training CRF using primitive features, we obtain the weights $\lambda_{r}^{t_o}$ for $r \in R$ and $t_o \in T$, which are very important to know which primitive features are more useful for prediction and need to be expanded. We use entropy to measure the usefulness of a primitive feature. We compute the probability of each tag $t_o$ for $r$:
\begin{equation} \label{eq:pt}
\begin{split}
p^r(t_o)=\frac{\exp(\lambda_r^{t_o})}{\sum_{l=1}^{\left\vert{T}\right\vert}\exp(\lambda_r^{t_l})} .
\end{split}
\end{equation}
Based on Equation \ref{eq:pt}, we compute the entropy for a primitive feature indexed by $r$:
\begin{equation} \label{eq:ent}
\begin{split}
H(r)=-\sum_{o=1}^{\left\vert{T}\right\vert} p^r(t_o)\text{log}p^r(t_o) .
\end{split}
\end{equation}
The intuition of using entropy is that a salient primitive feature should favor some tags over the others so it has low entropy. 
We select primitive features that attain the maximum probability for tag $t_o$ and have entropies below $\delta$ (We set $\delta=0.3$ for $\left\vert{T}\right\vert=2$): 
\begin{equation} \label{eq:rto}
\begin{split}
R^{t_o}=\{r|H(r)<\delta \wedge t_o=\operatorname*{arg\,max}_{t_l}p^r(t_l)\}.
\end{split}
\end{equation}
We obtain a set of primitive features indexed by $R^{t_o}$ and use it to generate $(k, v)$ for tag $t_o$ since each primitive feature can be interpreted as a $(k, v)$. We group the same $k$ under $R^{t_o}$ to form the set $K^{t_o}$ and use the associated $v$ as initial knowledge value.

\section{Knowledge-based Training}
We train KCRF using knowledge-based features in this section. A knowledge-based feature simply tells whether a feature found in an example with a specified knowledge type has some values found in the current knowledge base (or KB). We use $x_n$ to denote the feature vectors with knowledge-based features for the $n$-th word and use $d\in D$ to denote a knowledge-based feature indexed by $d$ in $x_n$. So $x_{n,d}=1$ indicates that the $d$-th feature is a knowledge-based feature and the $n$-th word has a knowledge with type $k_d$ and initial knowledge value $v$ found in \textit{KB}: 
\begin{equation} \label{eq:kb}
\begin{split}
x_{n, d}=\mathbb{I}\big( (k_d, v) \in \textit{KB} \big) .
\end{split}
\end{equation}
For example, if (\text{[DEP, nmod:with, VBZ]}, \text{``works''}) $\in$ \textit{KB}, the word ``phone'' has a dependency relation knowledge-based feature with type $k=\text{[DEP, nmod:with, VBZ]}$ and $v=\text{``works''}$ and current word knowledge-based feature $k=\text{[WORD]}$ and $v=\text{``phone''}$. We denote the trained KCRF as $c$ and its parameters $\lambda^c$. It predicts on $x$ and generates probabilities $p(y|x; \lambda^c)$ for $y \in \mathcal Y$.

\section{Knowledge Expansion}
We perform sequence labeling on a large amount of unlabeled reviews under the same category as the target entity to expand the \textit{KB} using $c$. We assume that target entities under the same category share similar knowledge. Here the key point is to ensure the quality of the expanded knowledge since it is very easy to have harmful knowledge from unlabeled reviews without human supervision. We aggregate knowledge from \emph{reliable prediction}s on those reviews. To obtain a reliable prediction for the $n$-th word, we marginalize over $y_{1:N}$ of other positions except $n$ as:
\begin{equation} \label{eq:mar}
\begin{split}
p(y_n|x; \lambda^c)=\sum_{y_1} \cdots \sum_{y_{n-1}} \sum_{y_{n+1}} \cdots \sum_{y_N} p(y_{1:N}|x; \lambda^c).
\end{split}
\end{equation}
Then if a tag $t_o$ attains the maximum probability that is larger than a threshold:  
$ \max_{t_o}\big(p(y_n=t_o|x; \lambda^c)\big)>\delta',$
we consider it as a reliable prediction for tag $t_o$ at position $n$. The knowledge-based features $k_d$ and potential knowledge values associated with such a reliable prediction are considered as candidate knowledge. We use $\textit{cKB}^{t_o}$ as the set of candidate knowledge for tag $t_o$. We further prune the knowledge since similar knowledge may appear in the knowledge base of another tag so this can make candidate knowledge from reliable predictions not reliable. 

\begin{algorithm}[t]
\DontPrintSemicolon
\caption{Knowledge Expansion}
\label{alg:ke}
\SetKwInOut{Input}{Input}
\SetKwInOut{Output}{Output}
\SetKwRepeat{Do}{do}{while}
\Input{$(c, \textit{KB})$, with $\textit{KB}=\{\textit{KB}^{t_o}|t_o \in T\}$, $U=\{u_1, ..., u_{\left\vert{U}\right\vert}\}$}
\Output{$(c, \textit{KB})$, with updated \textit{KB}}
\BlankLine
\BlankLine
\Do{$\textit{cKB} \neq \emptyset$}{
    transform each $u\in U$ into a sequence of knowledge-based feature vectors $x \in X$ using \textit{KB}\;\label{alg:trans}
    
    \For{$x \in X$}{
        use $(c, \textit{KB})$ to predict\;
        use Equation \ref{eq:mar} to compute $p(y_n|x; \lambda^c)$ for $n=1:N$\;\label{alg:mar}
        \For{$n=1, ..., N$}{
            \For{$t_o \in T$}{
                \If{$\max_{t_o}(p(y_n=t_o|x))>\delta'$}{\label{alg:if}
                    add associated $(k, v)$ to $\textit{cKB}^{t_o}$ for $k \in K^{t_o}$\;\label{alg:if2}
                }
            }
        }
    }
    \For{$t_o \in T$}{\label{alg:merge1}
        $\textit{cKB}^{t_o} \gets \textit{cKB}^{t_o}-\cup_{t_l\neq t_o} \textit{cKB}^{t_l} $ \;
        $\textit{KB}.\textit{KB}^{t_o} \gets \textit{KB}^{t_o}$ $\cup \textit{cKB}^{t_o}$ \;\label{alg:update}
    }\label{alg:merge2}
    $\textit{cKB} \gets \cup_{t_o} \textit{cKB}^{t_o}$\;
}
\Return{$(c, \textit{KB})$}
\end{algorithm}

Algorithm \ref{alg:ke} is to maintain high-quality knowledge during expansion. We use $U$ to denote a set of unlabeled sequences and we transform $u\in U$ to knowledge-based feature vectors $x\in X$ based on current knowledge base $\textit{KB}$ (line \ref{alg:trans}). We apply KCRF $c$ and current \textit{KB} on $x$ and get reliable prediction in line \ref{alg:if}. We add associated knowledge in line \ref{alg:if2}) and prune it to get reliable knowledge and update \textit{KB} in line \ref{alg:merge1}-\ref{alg:merge2}. The whole process will stop once no reliable knowledge is available. Note that during knowledge expansion, KCRF $c$ itself is never re-trained.

\section{Experimental Results}
\label{sec:exp}

\begin{table*}[!htp]
\centering
\scalebox{0.85}{
\begin{tabular}{ l | c c c | c c c | c c c | c c c }
\hline
\multirow{2}{*}{Product} & 
\multicolumn{3}{ |c| }{\textbf{CRF(-)DR}} & 
\multicolumn{3}{ |c| }{\textbf{CRF}} &
\multicolumn{3}{ |c| }{\textbf{CRF-Init}} &
\multicolumn{3}{ |c }{\textbf{KCRF}}\\
\cline{2-13}&
$\mathcal{P}$&$\mathcal{R}$&$\mathcal{F}_1$&
$\mathcal{P}$&$\mathcal{R}$&$\mathcal{F}_1$&
$\mathcal{P}$&$\mathcal{R}$&$\mathcal{F}_1$&
$\mathcal{P}$&$\mathcal{R}$&$\mathcal{F}_1$\\
\hline
Stylus  &  0.5 & 0.54 & 0.52 & 
0.75 & 0.50 & 0.60 & 
0.84 & 0.64 & 0.73 & 
0.66 & 0.84 & \textbf{0.74} \\
Micro SD Card  &  0.63 & 0.51 & 0.56 & 
0.89 & 0.44 & 0.59 & 
0.87 & 0.57 & 0.69 & 
0.77 & 0.70 & \textbf{0.74} \\
Mouse  &  0.54 & 0.37 & 0.44 & 
0.80 & 0.48 & 0.60 & 
0.75 & 0.53 & 0.62 & 
0.68 & 0.68 & \textbf{0.68} \\
Tablet Stand  &  0.58 & 0.43 & 0.49 & 
0.79 & 0.40 & 0.53 & 
0.85 & 0.46 & 0.60 & 
0.75 & 0.65 & \textbf{0.70} \\
\hline
Keyboard  &  0.54 & 0.46 & 0.5 & 
0.8 & 0.42 & 0.55 & 
0.8 & 0.34 & 0.48 & 
0.66 & 0.72 & \textbf{0.69} \\
Notebook Sleeve  &  0.69 & 0.38 & 0.49 & 
0.90 & 0.23 & 0.37 & 
0.91 & 0.23 & 0.37 & 
0.77 & 0.63 & \textbf{0.69} \\
Compact Flash  &  0.75 & 0.61 & 0.67 & 
0.92 & 0.46 & 0.62 & 
0.89 & 0.51 & 0.65 & 
0.82 & 0.73 & \textbf{0.77} \\
\hline
\end{tabular}
}
\caption{Comparison of different methods in precision, recall and F1-score}
\label{table:comparison}
\end{table*}

\subsection{Dataset}
We use the dataset\footnote{\url{https://www.cs.uic.edu/~hxu/}} from \cite{xu2016CER}, which includes 7 products. We take 50\% reviews of the first 4 products as the training data for all methods that require supervised training. The remaining reviews of the first 4 products (for in-domain test) and all reviews of the last 3 products (for out-of-domain test) are test data. 
Similar to \cite{xu2016CER}, we also randomly select 6000 unlabeled reviews for each category from \cite{McAPanLes15} and use them as unlabeled reviews to expand knowledge. 

\subsection{Compared Methods}
Since this paper proposes a supervised method on CER, we focus on the improvements of KCRF over CRF. We use CRFSuite\footnote{http://www.chokkan.org/software/crfsuite/} as the base implementation of CRF.\\
\textbf{CRF(-)DR}: This is a very basic CRF without dependency relations as features to show that dependency relations are useful features. We use the following features: current word, POS-tags, 4 nearby words and POS-tags on the left and right, number of digits and whether current word has slash/dash.\\
\textbf{CRF}: This is the baseline with dependency relations as features. It is also the same learner as in the pre-training step of KCRF.\\  
\textbf{CRF-Init}: This baseline uses the trained KCRF and initial KB directly on test data without knowledge expansion on unlabeled data.\\
\textbf{KCRF}: This is the proposed method that uses trained KCRF and initial KB to expand knowledge on unlabeled reviews under the same category as the target entity. We empirically set $\delta'=0.8$ as the precisions of most predictions are around 0.8. 

\subsection{Evaluation Methods}
We perform evaluation on each mention of complementary entities. We count true positive \textit{tp}, false positive \textit{fp} and false negative \textit{fn}. For each sentence, one recognized complementary entity that is contained in the annotated complementary entities for that sentence is considered as one count for the \textit{tp}; one recognized complementary entity that are not found contributes one count to the \textit{fp}; any annotated complementary entity that can not be recognized contributes one count to the \textit{fn}. Then we compute precision $\mathcal P$, recall $\mathcal R$ and F1-score $\mathcal F1$ based on \textit{tp}, \textit{fp} and \textit{fn}.

\subsection{Result Analysis}
From Table \ref{table:comparison}, we can see that KCRF performs well on F1-score. It significantly outperforms other methods on recall, which indicates that the expanded knowledge is helpful. CRF-Init performs better than CRF on most products, which indicates that knowledge-based features are better than primitive features in general. However, we notice that in order to get a higher recall, KCRF sacrifices its precision a lot. So how to further ensure that the expanded knowledge is of high quality to keep high precision is still an open problem. 

The performance of KCRF does not drop much for the last 3 products even though we do not have any training data for those products. This is because KCRF can utilize unlabeled data to expand knowledge about the last 3 products separately from the knowledge of the first 4 products. One intuitive example is that ``work'' can be a frequent general verb knowledge that exists in the training data for some verb related knowledge type. Then later KCRF expands such a verb to other domain-specific verbs like ``insert'' for Compact Flash that does not have training data.  

\section{Conclusion}
In this paper, we propose a supervised method for Complementary Entity Recognition (CER), called KCRF. KCRF can automatically identify knowledge-based features and expand knowledge as key-value pairs from plenty of unlabeled reviews. Experiments show that the expanded knowledge is useful in improving the performance of predictions, especially for products without training data.

\bibliography{emnlp2017}
\bibliographystyle{emnlp_natbib}

\end{document}